\pdfoutput=1
\documentclass[sigconf]{acmart}
\usepackage{graphicx}
\usepackage{amsmath}
\usepackage{booktabs}

\usepackage{times}
\usepackage{graphicx}
\usepackage{amsmath}

\usepackage{amssymb}
\usepackage{algorithm}
\usepackage{algpseudocode}
\usepackage{threeparttable}
\usepackage{multirow}
\usepackage{booktabs}
\usepackage{caption}
\usepackage{mathrsfs}
\usepackage{bbding}

\AtBeginDocument{%
  \providecommand\BibTeX{{%
    \normalfont B\kern-0.5em{\scshape i\kern-0.25em b}\kern-0.8em\TeX}}}

\makeatletter
\def\MT@is@composite#1#2\relax{%
	\ifx\\#2\\\else
	\expandafter\def\expandafter\MT@char\expandafter{\csname\expandafter
		\string\csname\MT@encoding\endcsname
		\MT@detokenize@n{#1}-\MT@detokenize@n{#2}\endcsname}%
	\ifx\UnicodeEncodingName\@undefined\else
	\expandafter\expandafter\expandafter\MT@is@uni@comp\MT@char\iffontchar\else\fi\relax
	\fi
	\expandafter\expandafter\expandafter\MT@is@letter\MT@char\relax\relax
	\ifnum\MT@char@ < \z@
	\ifMT@xunicode
	\edef\MT@char{\MT@exp@two@c\MT@strip@prefix\meaning\MT@char>\relax}%
	\expandafter\MT@exp@two@c\expandafter\MT@is@charx\expandafter
	\MT@char\MT@charxstring\relax\relax\relax\relax\relax
	\fi
	\fi
	\fi
}
\def\MT@is@uni@comp#1\iffontchar#2\else#3\fi\relax{%
	\ifx\\#2\\\else\edef\MT@char{\iffontchar#2\fi}\fi
}
\makeatother

\setcopyright{acmcopyright}

\begin{document}

\title{Dilated Context Integrated Network with Cross-Modal Consensus for Temporal Emotion Localization in Videos}

%
%
%
%
%
%
%
%
%
%
%
%
%
%
%
%
%

\author{Juncheng Li$~\textsuperscript{\rm 1*}$ Junlin Xie$~\textsuperscript{\rm 1*}$ Linchao Zhu$~\textsuperscript{\rm 2}$ Long Qian$~\textsuperscript{\rm 1}$ Siliang Tang$~\textsuperscript{\rm 1}$ Wenqiao Zhang$~\textsuperscript{\rm 3}\dag$}
	
\author{Haochen Shi$~\textsuperscript{\rm 4}$ Shengyu Zhang$~\textsuperscript{\rm 1}$  Longhui Wei$~\textsuperscript{\rm 5}$ Qi Tian$~\textsuperscript{\rm 5}$ Yueting Zhuang$~\textsuperscript{\rm 1}$}
\affiliation{%
	\institution{$~\textsuperscript{\rm 1}$ Zhejiang University $~\textsuperscript{\rm 2}$ University of Technology Sydney $~\textsuperscript{\rm 3}$ National University of Singapore}
	 \country{}
}
\affiliation{%
	\institution{$~\textsuperscript{\rm 4}$ Universit\'{e} de Montr\'{e}al $~\textsuperscript{\rm 5}$Huawei Cloud}
	\country{}
}
\affiliation{%
	\institution{\{junchengli, junlinxie, qianlong0926, siliang, sy\_zhang, yzhuang\}@zju.edu.cn}
	\country{}
}
\affiliation{%
	\institution{linchao.zhu@uts.edu.au\quad wenqiao@nus.edu.sg \quad haochen.shi@umontreal.ca} 
	\country{}
}
\affiliation{%
	\institution{weilh2568@gmail.com\quad tian.qi1@huawei.com}
	\country{}
}

\thanks{* Equal Contribution.}
\thanks{\dag \  Corresponding Author.}

\renewcommand{\shortauthors}{Juncheng Li et al.}



\begin{abstract}
	Understanding human emotions is a crucial ability for intelligent robots to provide better human-robot interactions. The existing works are limited to trimmed video-level emotion classification, failing to locate the temporal window corresponding to the emotion. In this paper, we introduce a new task, named Temporal Emotion Localization in videos~(TEL), which aims to detect human emotions and localize their corresponding temporal boundaries in untrimmed videos with aligned subtitles. TEL presents three unique challenges compared to temporal action localization: 1) The emotions have extremely varied temporal dynamics; 2) The emotion cues are embedded in both appearances and complex plots; 3) The fine-grained temporal annotations are complicated and labor-intensive. To address the first two challenges, we propose a novel dilated context integrated network with a coarse-fine two-stream architecture. The coarse stream captures varied temporal dynamics by modeling multi-granularity temporal contexts. The fine stream achieves complex plots understanding by reasoning the dependency between the multi-granularity temporal contexts from the coarse stream and adaptively integrates them into fine-grained video segment features. To address the third challenge, we introduce a cross-modal consensus learning paradigm, which leverages the inherent semantic consensus between the aligned video and subtitle to achieve weakly-supervised learning. We contribute a new testing set with 3,000 manually-annotated temporal boundaries so that future research on the TEL problem can be quantitatively evaluated. Extensive experiments show the effectiveness of our approach on temporal emotion localization. The repository of this work is at \url{https://github.com/YYJMJC/Temporal-Emotion-Localization-in-Videos}.
\end{abstract}

\copyrightyear{2022} 
\acmYear{2022} 
\acmConference[MM '22]{Proceedings of the 30th ACM International Conference on Multimedia}{October 10--14, 2022}{Lisboa, Portugal} \acmBooktitle{Proceedings of the 30th ACM International Conference on Multimedia (MM '22), October 10--14, 2022, Lisboa, Portugal}
\acmDOI{10.1145/3503161.3547886} 
\acmISBN{978-1-4503-9203-7/22/10}




\maketitle
\section{Introduction}

\begin{figure}[!t]
	\centering
	\includegraphics[width=\linewidth]{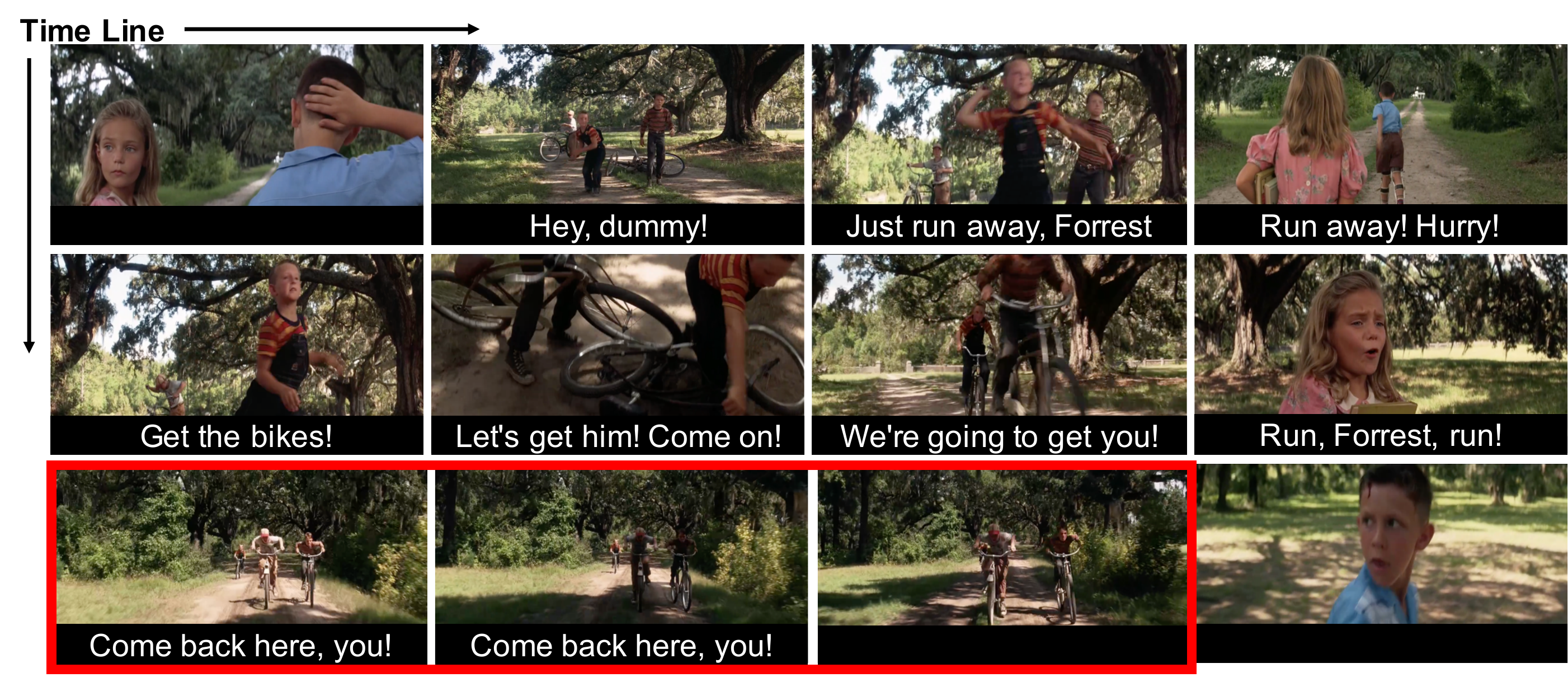}
	\vspace{-0.6cm}
	\caption{We show some key segments and their aligned subtitles. The segments in the red box are the target segments that temporally occur \textsl{Anger} emotion.}
	\label{demo}
	\vspace{-0.4cm}
\end{figure}

Humans are social creatures and thrive on empathy. We can easily put ourselves in other's situations and make decisions based on the inferences of other's internal states~(\textsl{i.e.} emotional states). Recent studies~\cite{barbey2014distributed, herbet2014inferring} on neuroscience confirm that emotional intelligence and cognitive intelligence share many neural systems for integrating cognitive, social, and affective processes. Therefore, understanding emotions is crucial for achieving high-level intelligence. In practice, this ability helps social chatbots and personal assistants to better understand the mood and motivations of people, so they can better interact with people.


Emotion understanding has long been studied in computer vision. Previous studies~\cite{fabian2016emotionet, wei2020learning, ali2017high, pilli2020predicting} mainly focus on recognizing emotion through facial expression in static images. As movies provide diverse social situations that are closer to our daily life, some recent studies~\cite{mittal2021affect2mm, hazer2015emotion, mcduff2014predicting} have been proposed to classify emotions from movies. However, they are limited to video-level classification, failing to identify the corresponding temporal boundaries of the emotions, which is essential in practical application. To break through the above limitation, we propose a novel task of Temporal Emotion Localization in videos~(TEL), which aims to predict emotions and their corresponding start and end timestamps in untrimmed videos with aligned subtitles.

Compared with conventional temporal action localization (TAL) \cite{caba2015activitynet}, TEL presents some unique challenges. First, the emotions have extremely varied temporal dynamics. Such fine-grained emotions could appear in arbitrary frames and last for varied durations, causing a great challenge for temporal localizing. For example, the peace emotion may exist for a long time but the surprise emotion may happen quickly. Even the same emotion may have extremely varied durations in different situations. Secondly, unlike existing action localization, where the actions have more consistent visual patterns and only rely on a single modality~(video) as the context, in TEL, the emotion cues are embedded in both appearances and complex plots with multi-modality context. In TEL, the videos paired with subtitles come from movies, which contain diverse event dynamics and character interactions, and the emotions are more ambiguous and have higher inter-class similarity than action classes. To discriminate very similar emotions, the model needs to achieve in-depth comprehension of complex plots by jointly reasoning over multi-modal and multi-granularity temporal context. As shown in Figure \ref{demo}, when we look only at the target segments in the red box, we can guess that these boys are playing and feeling \textsl{Happiness} and \textsl{Engagement}, but it is hard to identify more specific cues for their emotions. When we further see the corresponding subtitles, we may infer that they are chasing and feeling \textsl{Excitement}. However, only when we consider the whole context that they are bullying a boy and chasing him, can we say they are probably feeling anger. Thirdly, annotating fine-grained temporal boundaries of emotions in videos is complicated and labor-intensive. Thus, a weakly-supervised algorithm for TEL is more widely available.

In this paper, we propose a novel dilated context integrated network with cross-modal consensus learning to address the aforementioned challenges. For the first two challenges,  we introduce a Dilated Context Integrated Network~(DCIN) that adaptively models multi-granularity temporal dynamics to achieve in-depth understanding of complex plots. Specifically, DCIN models temporal context in a coarse-fine two-stream architecture. The coarse stream models multiple abstract-level of context in a hierarchical structure to capture varied temporal dynamics. The fine stream reasons the temporal dependency between the multi-granularity temporal context and adaptively integrates them into fine-grained video segment features by joint reasoning over video and subtitle, which achieves in-depth plots understanding. Furthermore, we present a context-sensitive constraint to encourage the DCIN to learn more discriminative context that can help to determine the emotion.


For weakly-supervised learning, we propose a Cross-Modal Consensus Learning~(CCL) paradigm by leveraging the inherent semantic consensus between the aligned video and subtitle. The intuition behind this is that when we see the subtitle ``thanks for your delicate gift" we can easily infer the visual situation that somebody is happy and vice versa. If we see a video segment of somebody happy to accept a gift, we may infer some subtitles expressing his happiness. Concretely, given the ground-truth emotion label without temporal annotation, the model first identifies the most relevant video segment and then uses its temporally co-occurring subtitle to predict the most possible emotion. We train our model such that the predicted emotion based on subtitle is consistent with the original emotion for retrieving the most relevant video segment. Further, our empirical experiments indicate that sometimes the alignment between video and subtitle is noisy as the subtitle might refer to previous or forthcoming visual events. Therefore, to alleviate the misalignment noise, we present a temporal alignment relaxation strategy, which enables the model to dynamically learn the alignment from CCL paradigm.

To facilitate research of the TEL task, we contribute a testing set by manually annotating the temporal boundaries of 3000 samples on the MovieGraph dataset~\cite{vicol2018moviegraphs}. In summary, our contributions are:

\begin{itemize}
	\item We define a new task, Temporal Emotion Localization in videos~(TEL), to provide a new benchmark for the research on emotion understanding and video-and-language reasoning.
	\item We propose a novel Dilated Context Integrated Network (DCIN) that adaptively integrates multi-granularity temporal context in a coarse-fine architecture, of which we introduce a context-sensitive constraint to enforce the integrated context to be more emotion-discriminative.
	\item We propose a novel Cross-Modal Consensus Learning~(CCL) for weakly-supervised TEL, which utilizes the cross-modal semantic consensus between video and subtitle.
	\item The proposed framework outperforms the baselines by a large margin and can transfer to several video-and-language tasks.
	
\end{itemize}

\section{Related Work}
\noindent
\textbf{Emotion Understanding.} Emotion understanding has long been studied in computer vision. Existing researchers mainly focus on recognizing emotion through facial expressions. Quiroz \textsl{et al.}~\cite{fabian2016emotionet} propose a large dataset of one million images of facial expressions of emotion in the wild. Wei \textsl{et al.}~\cite{wei2020learning} perform emotion recognition by learning a feature extraction network on StockEmotion, which has more than a million images. 
Instead of recognizing emotions only based on facial expressions, there has been a growing interest in dynamically modeling emotions over time. Movies serve as an appropriate testbed of emotion understanding, as they contain multimodal context and diverse human emotions in a variety of situations. Several works~\cite{mittal2021affect2mm, schaefer2010assessing, hazer2015emotion, mcduff2014predicting} have been proposed to classify emotions from movie clips. Although they have achieved promising performance, they still remain in the emotion classification on the whole video, lacking transparency to tell which segments of the video the emotion appears in. In contrast, we further explore localizing the start and end points of emotions in untrimmed videos, which is more challenging and crucial for understanding human emotions in real-world situations (\textsl{e.g.}, personal assistants, e-commerce~\cite{DBLP:conf/mm/ZhangTYZKLZYW2020}).

\noindent
\textbf{Action Localization.} Action localization~\cite{caba2015activitynet} aims to predict actions and corresponding start and end timestamps in videos. In general, existing supervised methods can be categorized into top-down and bottom-up frameworks. The top-down methods~\cite{buch2017sst, heilbron2016fast, escorcia2016daps, zeng2019graph, chao2018rethinking} first extract a set of candidate proposals and refine
them to achieve the final temporal boundaries. The bottom-up methods~\cite{buch2019end, lin2019bmn, lin2018bsn, liu2019multi, long2019gaussian} directly predict frame-level or snippet-level scores and then combine the individual scores to generate the final temporal boundaries. Since supervised action localization requires labor-intensive frame-level annotations, weakly-supervised action localization~\cite{wang2017untrimmednets, nguyen2018weakly, narayan20193c, ma2021weakly}  has received increasing attention.

\noindent
\textbf{Video-and-Language Understanding.} The advent of deep learning~\cite{lecun2015deep, li2020ib, kong2022attribute, guo2021semi, guo2022collaborative} promotes the prosperity of computer vision~\cite{li2020multi, DBLP:conf/mm/ZhangTYZKLZYW20, zhang2022boostmis} and vision-and-language~\cite{li2020unsupervised, li2019walking, zhang2022magic, li2020topic, DBLP:conf/mm/ZhangJWKZZYYW20, zhang2021consensus, zhang2020photo, https://doi.org/10.48550/arxiv.2207.04211}. With the flourishing development of large-scale video datasets~\cite{krishna2017dense, abu2016youtube, kay2017kinetics}, several video-and-language understanding tasks~\cite{li2021adaptive, DBLP:conf/kdd/ZhangTZYKJZYW20, DBLP:conf/mm/ZhangTYZKLZYW20} have received increasing attention, such as temporal sentence grounding~\cite{gao2017tall, li2022compositional, li2022end, li2022hero}, video question answering~\cite{lei2018tvqa, tapaswi2016movieqa, zhang2019frame}, and video captioning~\cite{zhang2020relational}. 
These tasks mainly focus on identifying explicit visual cues (\textsl{e.g.,} objects, actions, characters), which are mainly embedded in obvious visual appearances. TEL differs as it requires more sophisticated reasoning skills, such as understanding complex plots, reasoning character relationships, and inferring human's internal states. These abilities can facilitate more sensible human-robot interactions based on a better comprehension of human emotions.

\vspace{-0.3cm}
\section{Method}


\noindent
\textbf{Problem Formulation.} Given an untrimmed video $V$ paired with subtitle $S$, we aim to detect emotions in the video and locate their corresponding segments. As an untrimmed video might involve multiple emotions, we formulate the problem as a multi-label detection problem. For weakly-supervised setting, only the video-level emotion labels are available, without any temporal boundary annotations. The video is represented as $V = \{v_i\}^T_{i=1}$ segment-by-segment, and the subtitle is represented as $S = \{s_i\}^T_{i=1}$ sentence-by-sentence, where $s_i$ represents the subtitle sentence that is temporally co-occurring with $v_i$. We obtain $v_i \in R^{1 \times d}$ by max-pooling over the I3D~\cite{carreira2017quo} features of frames within the segment. $s_i \in R^{1 \times d}$ is the sentence embedding.

\subsection{Dilated Context Integrated Network}\label{s3.1}
As aforementioned, the key factor for fine-grained emotion localization is the multi-modality and multi-granularity context modeling. Existing action localization methods mainly use RNN~\cite{hochreiter1997long}, 3D~CNN~\cite{tran2015learning,carreira2017quo}, or Transformer~\cite{vaswani2017attention} to recognize specific visual patterns of actions. However, they are unsuitable for the complex multi-modality and multi-granularity context modeling. For RNN-based methods, they do not capture non-sequential temporal dependencies effectively. For 3D~CNN-based methods, they suffer from limited temporal receptive field. For Transformer-based methods, such fully-connected structures may cause the fine-grained local context to be overwhelmed by unimportant information.

\begin{figure}[!t]
	\centering
	\includegraphics[width=\linewidth]{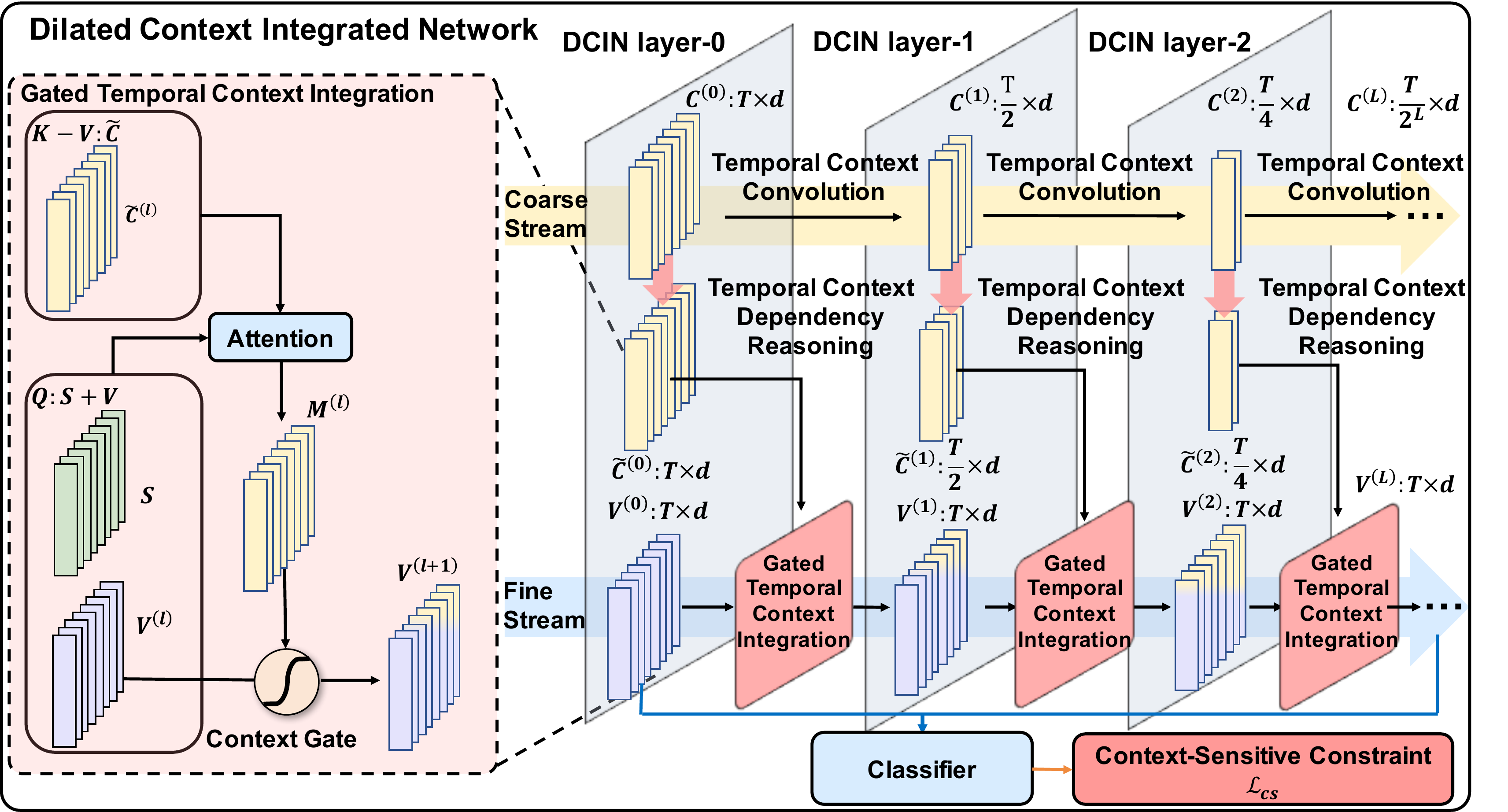}
	\vspace{-0.6cm}
	\caption{Overview of our Dilated Context Integrated Network.} 
	\label{DCIN}
	\vspace{-0.4cm}
\end{figure}

Differently, we present the Dilated Context Integrated Network (DCIN) that adaptively integrates multiple abstract-level of context into fine segment representations in a hierarchy. As shown in Figure \ref{DCIN}, DCIN processes the information in a two-stream architecture. The coarse stream learns to model multi-granularity temporal context in a hierarchy. The fine stream reasons the temporal dependency among the temporal context and gradually fuses the multi-granularity context from the coarse stream with the fine segment representations. To avoid unnecessary and redundant context, we propose a novel gated temporal context integration module to dynamically integrate informative context by joint reasoning over video and subtitle. Furthermore, we introduce the context-sensitive constraint to encourage the model to learn more discriminative context that can help to determine the emotion. Concretely, each DCIN layer consists of: 1)~Temporal Context Convolution, 2)~Temporal Context Dependency Reasoning, and 3)~Gated Temporal Context Integration. 

\noindent
\textbf{Temporal Context Convolution.} For the fine-grained emotion localization, the model must discriminate very similar emotions through video context, which may have various durations and scales. Thus, we use temporal context convolution to generate multi-granularity temporal context. Specifically, we use 1D temporal convolution operation with $stride = 2$ to halve the temporal dimension of the context at each layer. We define the context produced by the Coarse stream at layer $l$ as $C^{(l)}$, where $C^{(0)} = V$. Given $C^{(l-1)} \in R^{T_{l-1} \times d}$ from the previous layer, we compute $C^{(l)}\in R^{T_{l} \times d}$ ($T_{l} = T_{l-1}/2$) as:

\begin{equation}
C^{(l)} = f(W_1 \ast C^{(l-1)} + b_1)
\end{equation}
where $f(\cdot)$ is the activation function, $\ast$ is the convolution operator, and $W_1$ is the 1D convolution filters. As a consequence,  we get $C^{(l)}$ that contains increasing levels of semantic meaning and higher temporal resolution context.

\noindent
\textbf{Temporal Context Dependency Reasoning.} The video clips are collected from movies, which contain complex event dynamics and diverse character interactions across multiple segments. Therefore, we develop the temporal context dependency reasoning to capture the non-local temporal structure of context. Concretely, we adopt graph convolution on the context features $C^{(l)} = \{c_i^{l}\}_{i=1}^{T_l}$ as:

\begin{equation}
\begin{aligned}
\tilde{c}_i^l\!=\!c_i^l\!+\!\sum_{j}^{T_l} \alpha_{ij}^{tcdr} \cdot (W_2c_j^l), \quad \alpha_{ij}^{tcdr}\!=\!\frac{exp({c_i^l}^T \cdot c_j^l)}{\sum_{j}^{T_l} exp({c_i^l}^T \cdot c_j^l)}
\end{aligned}
\end{equation}

\noindent
where $W_2 \in R^{d \times d}$ is the learnable projection matrix, $\alpha_{ij}^{tcdr}$ is the semantic coefficient between context node $c_i^l$ and $c_j^l$.

\noindent
\textbf{Gated Temporal Context Integration.} After obtaining the non-local enhanced context features $\tilde{C}^{(l)}$, we propose a gated temporal context integration module to adaptively integrate context  $\tilde{C}^{(l)}$ into segment features $V^{(l-1)}$ at layer $l-1$ ($V^{(0)} = V$) to get $V^{(l)}$. Considering the complementary nature of video and subtitle, we design a cross-modal context filtering mechanism that utilizes the aligned subtitle feature $s_i$ of $v_i$ to select relevant context. The aggregated context information for segment $v^{l-1}_i$ is computed as:

\begin{equation}
m_i\!=\!\sum_{j}^{T_l} \alpha_{ij}^{gtci} \cdot \tilde{c}_j^l, \quad \alpha_{ij}^{gtci}\!=\!\frac{exp({v^{l-1}_i}^T \cdot \tilde{c}_j^l\!+\!{s_i}^T \cdot \tilde{c}_j^l)}{\sum_j^{T_l} exp({v^{l-1}_i}^T \cdot \tilde{c}_j^l\!+\!{s_i}^T \cdot \tilde{c}_j^l)}
\end{equation}

\noindent
where the cross-modal feature $s_i$ helps to reassign the semantic coefficient $\alpha_{ij}^{gtci}$. Subsequently, we build the context gate $g_i$ that controls the flow of aggregated context $m_i$ to $v^{l-1}_i$, and update the fine segment representations:

\begin{equation}
g_i =  \sigma(W_3[v_i^{l-1}, m_i] + b_3), \quad v^l_i = (1 - g_i) \odot v^{l-1}_i + g_i \odot m_i
\end{equation}

\noindent
As a consequence, we obtain segment features $V^{(l)}$ that integrate context features $\tilde{C}^{(l)}$. By performing $L$ DCIN layers, we gradually integrate increasing levels of temporal context into fine segment features and learn final context-aware segment features $V^{(L)} = \{v_i^L\}_{i=1}^T$.

\noindent
\textbf{Context-Sensitive Constraint.} Fine-grained emotion localization requires the model to discriminate similar emotions such as embarrassment and disquietment. For human beings, they will turn to the rich context that implicates multi-granularity event semantics to disambiguate the emotions. Motivated by this insight, we propose a novel context-sensitive constraint to encourage the model to learn more emotion-discriminative context. Intuitively, we can use the original segment representations $V^{(0)}$  and the context-aware segment representations $V^{(L)}$ to predict the emotion class probability distributions, respectively. If the model is sensitive to the context, the predicted probability will change greatly after integrating the multi-granularity context. 
Thus, the distance between the two predicted probability distributions should be far. Specifically, given the original segment representations $V^{(0)} = \{v_i^0\}^T_{i=1}$ and the context-aware segment representations $V^{(L)} = \{v_i^L\}_{i=1}^T$, we first compute the  probability of each emotion class for each $v_i^0$ and $v_i^L$ as:

\begin{equation}
P(E|v_i^0)\!=\! f(W_4v_i^0 + b_4),\ P(E|v_i^L)\!=\!f(W_4v_i^L + b_4)
\end{equation}

\noindent
where $P(E|\cdot) \in  R^{N \times 1}$ and $N$ is the number of emotion classes. Next, we adopt Euclidean distance to measure the context sensitivity for a pair of $v_i^0$ and $v_i^L$ as:

\begin{equation}
d(v_i^0, v_i^L) = || P(E|v_i^L) - P(E|v_i^0) ||
\end{equation}

\noindent
Then the context-sensitive constraint loss is formulated as:

\begin{equation}
\mathcal{L}_{cs} = \sum_{i}^{T} max(0, \Delta - d(v_i^0, v_i^L))
\end{equation}

\noindent
where $\Delta$ is the margin hyper-parameter. By minimizing $\mathcal{L}_{cs}$, we enlarge the distance between the two distributions, encouraging the model to be more sensitive to the context.

\begin{figure}[!t]
	\centering
	\includegraphics[width=\linewidth]{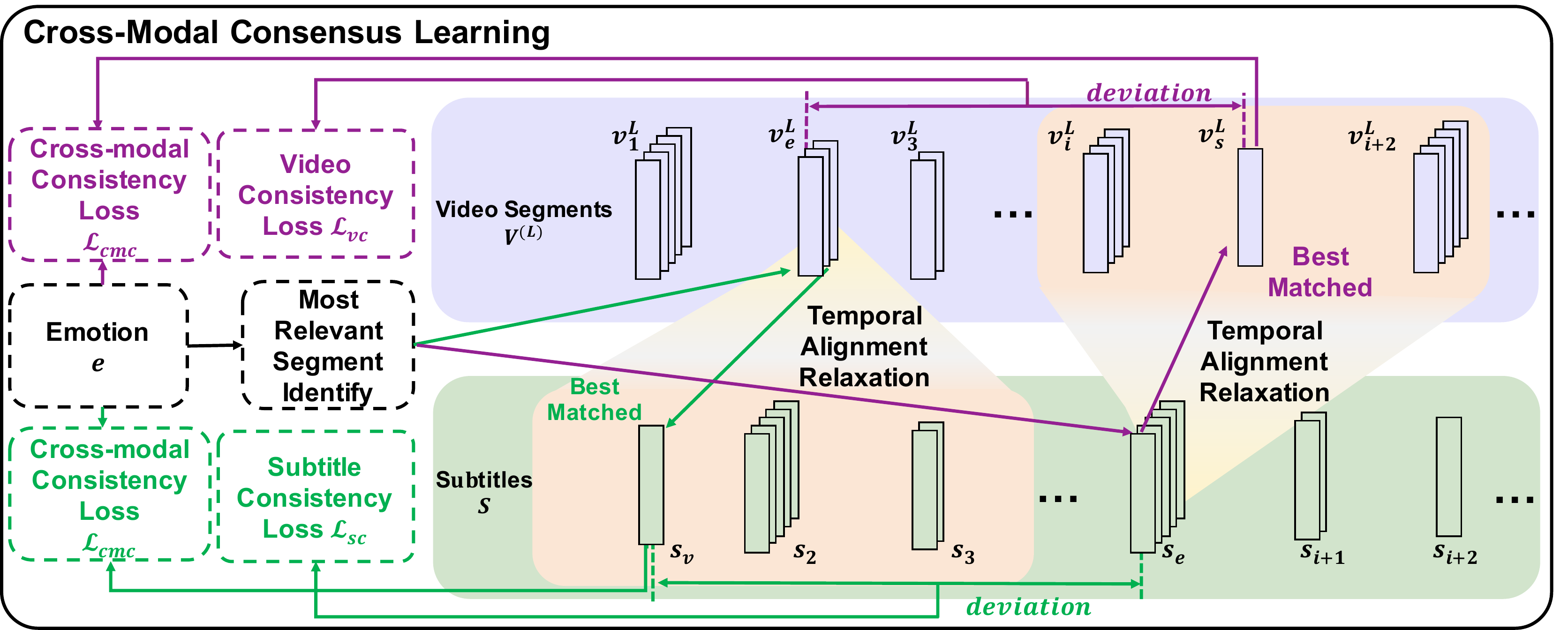}
	\vspace{-0.6cm}
	\caption{The framework of Cross-Modal Consensus Learning. }
	\label{CCL}
	\vspace{-0.3cm}
\end{figure}

\subsection{Cross-Modal Consensus Learning}\label{3.2}
Imagining seeing the subtitle ``thanks for your delicate gift", we will infer the visual situation that somebody is happy and vice versa. Inspired by this observation, we propose a cross-modal consensus learning~(CCL) paradigm by leveraging the semantic consensus between the aligned video and subtitle. As shown in Figure \ref{CCL}, given the ground-truth emotion label $e$ without temporal annotation, the model first identifies the most relevant video segment. The model then uses its temporally aligned  subtitle sentence to predict the most possible emotion. The visual and linguistic modalities are semantically consistent only if the predicted emotion label is the same as the ground-truth. 

Specifically, given the ground-truth emotion label $e$, we first identify the most relevant segment $v_e^L$ as:
\begin{equation}
v_e^L = \mathop{argmax}\limits_{v_i^L} P(e|v_i^L)
\end{equation}

\noindent
Then, we retrieve the subtitle sentence $s_v$ that is temporally aligned with $v_e$~(following, we omit the superscript $L$ for simplicity). Next, we compute the emotion class probability distribution based on $s_v$ as $P(E|s_v)$, and maintain the consensus of the score distributions based on $v_e$ and $s_v$ as:
\begin{equation}
\mathcal{L}_{cmc} = -\sum_{k}^N P(e_k|v_e)logP(e_k|s_v)
\end{equation}

\noindent
Where $N$ is the number of emotion classes, and $\mathcal{L}_{cmc}$ is the cross-modal consensus loss. Here we use $P(E|v_e)$ as the pseudo labels, and minimize the cross-entropy loss between them to encourage the semantic consensus. Besides from emotion to video to subtitle, we can also start from emotion to subtitle to video. Let $s_e$ denote the most relevant subtitle sentence and $v_s$ is the aligned video segment. The overall cross-modal consensus loss is:

\begin{equation} \label{e10}
\mathcal{L}_{cmc} = -\sum_{k}^N [P(e_k|v_e)logP(e_k|s_v) + P(e_k|s_e)logP(e_k|v_s)]
\end{equation}

Ideally, the most relevant segment $v_e$ should be the same as the segment $v_s$ retrieved  from the subtitle side. Thus, we penalize deviation between $v_e$ and $v_s$, which encourages the semantic consensus on video. The video consensus loss is defined as:

\begin{equation}
\mathcal{L}_{vc} = ||idx(v_e) - idx(v_s)||^2
\end{equation}

\noindent
where $idx(\cdot)$ is the segment index. In a similar manner, we can define the subtitle consensus loss as:

\begin{equation}  \label{e12}
\mathcal{L}_{sc} = ||idx(s_e) - idx(s_v)||^2
\end{equation}

\noindent
\textbf{Temporal Alignment Relaxation.} We observe that sometimes people might refer to previous or forthcoming visual events, so the temporal alignment between segment and subtitle may be noisy. In this regard, to alleviate the misalignment noise, we relax the hard temporal alignment constraint and encourage the model to dynamically learn the alignment from our cross-modal consensus learning paradigm. Concretely, for segment $v_e$, we first retrieve its temporal aligned subtitle sentence $s_v$. We then take the $Q$ sentences closest to $s_v$ in time  as the candidate set $Q(s_v)$. Next, we compute the semantic alignment score  between $v_e$ and  $s_q \in Q(s_v)$:
\begin{equation}
score(v_e, s_q) = cos(v_e, s_q) - \beta ||idx(v_e) - idx(s_q)||
\end{equation}

\noindent
where $cos(\cdot)$ is the cosine similarity and the second term is the index distance. Finally, we select the best match subtitle sentence as $s_v^*$:

\begin{equation}
s_v^* = \mathop{argmax}\limits_{s_q \in Q(s_v)} score(v_e, s_q)
\end{equation}

\noindent
And we can obtain the $v_s^*$ in a similar manner. Finally, We use the $s_v^*$ and $v_s^*$ to compute the losses in Equation \ref{e10} - \ref{e12}.

To differentiate through the cycle, previous methods are usually implemented as soft retrievals~(also viewed as attention mechanism). Differently, we implement the $argmax$ operation as a ``mask'' matrix that keeps track of where the maximum of the matrix is. And we empirically observe better performance on the ``mask" version.



\begin{figure}[!t]
	\centering
	\includegraphics[width=\linewidth]{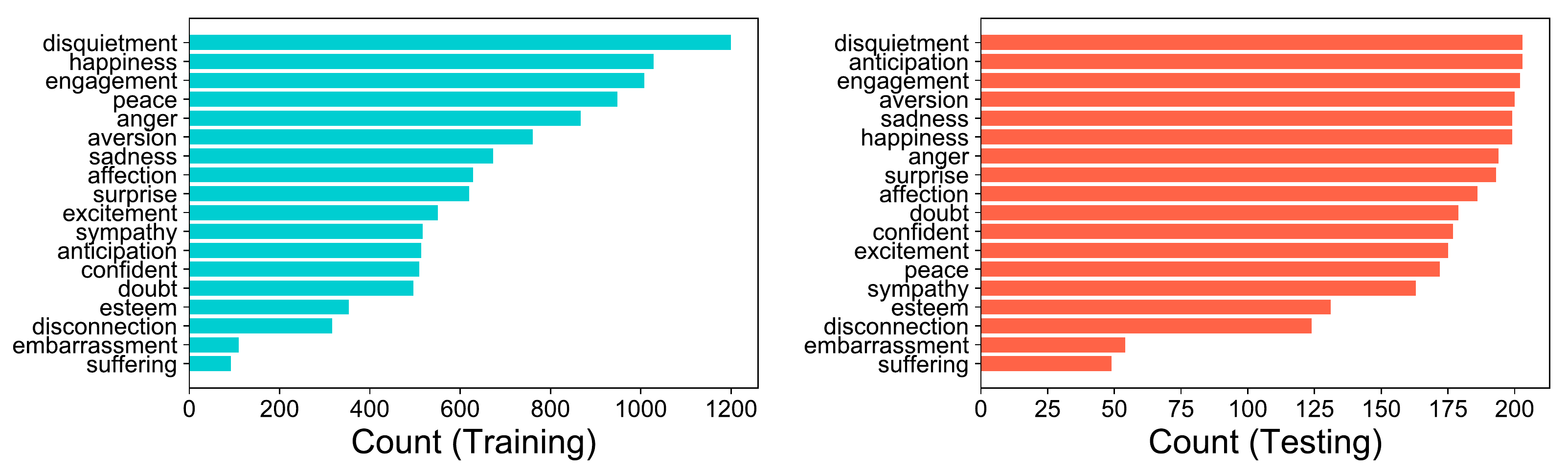}
	\vspace{-0.8cm}
	\caption{The distribution of 18 emotion classes.}
	\label{statistics}
	\vspace{-0.7cm}
\end{figure}

\vspace{-0.1cm}
\subsection{Training and Inference}\label{3.3}

\noindent
\textbf{Training.} The final training loss for the overall model is:
\begin{equation}
\mathcal{L} = \lambda_1 \mathcal{L} _{cs} + \lambda_2 \mathcal{L} _{cmc} + \lambda_3 \mathcal{L} _{vc} + \lambda_4 \mathcal{L} _{sc}
\end{equation}

\noindent
\textbf{Inference.} Given the above $V^{(L)}$ and $S$, the final emotion-segment matching score $m_i^e$ is defined as:

\begin{equation}
m_i^e = \frac{1}{2} ( P(e|v_i^L) + P(e|s_i) )
\end{equation}

\noindent
where $m_i^e \in R$ is the final matching score between segment $i$ and emotion label $e$, and $M_i = \{m_i^e\}^N_{e=1} \in R^{N \times 1}$ is the matching score of segment $i$. We first compute $\{M_i\}^T_{i=1}$ for segment sequence. Then, the emotions where the matching score is above threshold $m_i^e > \gamma_1$ are considered selected. Next, for each selected emotion $e^*$, we choose the video segment $v^*$ that has the highest matching score with $e^*$. Finally, we consider the segments adjacent to $v^*$. If their matching scores are above threshold $m_{\cdot}^{e^*} > \gamma_2$, we group them iteratively to form the final predicted temporal boundaries for $e^*$.

\section{Dataset}

\subsection{Fine-Grained Emotion Category Generation.} 
Existing datasets for emotion recognition are mainly based on still images and classify emotions according to 6 categories. In this work, we introduce a novel task that aims to localize fine-grained emotion in videos. Although several video-based emotion recognition datasets have been proposed, they mainly focus on single-person narratives recorded in controlled lab settings. In contrast, the recently released MovieGraphs dataset~\cite{vicol2018moviegraphs} contains rich real-world situations, diverse character interactions, and fine-grained emotion annotations. Thus, we evaluate our approach on it. MovieGraphs dataset provides detailed graph-based annotations of social situations for 7637 clips in 51 movies. The dataset was collected and manually annotated using crowd-sourcing methods. The emotion labels are represented as attribute nodes of actors. We extract 239 available emotion labels from all clips and group them into 18 discrete emotion classes. Specifically, we first use word connections (synonyms, relevance, affiliations) and the inter-dependence of a group of words (psychological research and affective computing) \cite{kosti2017emotic, fernandez2010psicologia} to form word-groupings. Then, we perform multiple iterations and cross-referencing with dictionaries and research in affective computing. Finally, we obtain the 18 emotion categories. The details on the definition of each emotion category and the grouped emotions in each category are provided in supplementary materials.

\vspace{-0.1cm}
\subsection{Dataset Annotation.}
We first split and clean the emotion localization samples. As only a few emotion labels in the MovieGraphs dataset have temporal annotations, we develop an annotation tool and ask human annotators to provide the temporal boundaries of emotions in video clips. Before officially annotating, we ask workers to annotate the same set of clips according to provided instructions and examples. For each annotation, we compute its average overlap with the annotations from other workers. If the average overlap is lower than a threshold, we will disregard the annotation. We choose the temporal intersection of consistent annotations as ground-truth. We also manually check the annotations that do not meet the consistency. Overall, 61\% of emotions are annotated by at least three workers, and 83\% of emotions are annotated by at least two workers. Finally, we take the annotated samples as the testing data. 

\begin{figure}[!t]
	\centering
	\includegraphics[width=\linewidth]{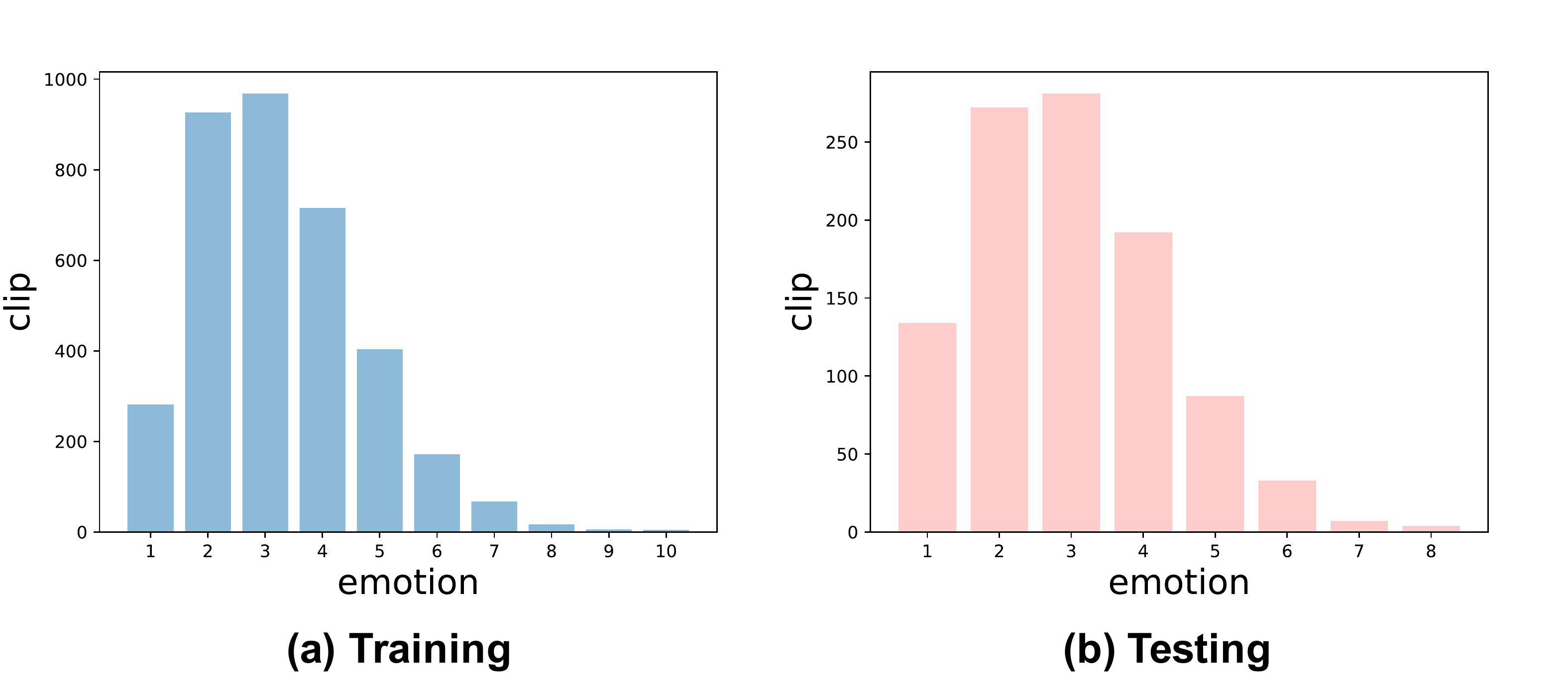}
	\vspace{-0.8cm}
	\caption{The distribution of the number of emotions per clip.}
	\label{emo_per_clip}
	\vspace{-0.6cm}
\end{figure}

\vspace{-0.1cm}
\subsection{Dataset Statistics}
The average duration of video clips is 44.28s and the average number of emotion labels is 3.6 per clip. We re-split the dataset into training~(39 movies) and testing~(12 movies). The training set consists of 11193 emotions, and the testing set consists of 3003 emotions with temporal annotations. Because some emotion classes are relatively rare in daily life, the distribution of emotion classes is not completely balanced. To comprehensively evaluate the model performance on all classes, we attempt to make the distribution of testing data relatively balanced. The distribution of 18 emotion types is illustrated in Figure \ref{statistics}. We also propose the number of emotions per clip over all training and testing movies in Figure \ref{emo_per_clip}.

\begin{table}[!t]
		\begin{threeparttable}
			\begin{tabular}{ lll cccc}
				\toprule
				\multicolumn{3}{l}{Method}
				
				&R@0.5 &R@0.7   &mAP  &mIoU\\				
				\hline    
				\multicolumn{3}{l}{Random}    &0.17 &0.07  &0.19     &0.15    \\
				\hline
				\multicolumn{3}{l}{Subtitle-Only}    &5.84     &3.10    &7.49     &5.71     \\
				\hline
				\multicolumn{3}{l}{UNets w/o subtitle} &1.86  &0.50 &2.36  &1.85 \\
				\multicolumn{3}{l}{UNets~\cite{wang2017untrimmednets}}  &7.06 &2.83  &8.38 &6.63 \\
				\multicolumn{3}{l}{3C-Net w/o subtitle}  &3.69     &1.30  &6.59 &5.09   \\
				\multicolumn{3}{l}{3C-Net~\cite{narayan20193c}}  &7.63   &2.40 &11.42 &8.95   \\
				\multicolumn{3}{l}{ASL w/o subtitle}  &4.80  &1.73  &9.25 &7.29   \\
				\multicolumn{3}{l}{ASL~\cite{ma2021weakly}}    &9.56   &3.13  &14.85  &11.35  \\
				\hline
				\multicolumn{3}{l}{XML w/o subtitle} &7.26  &3.07  &9.20  &7.81      \\
				\multicolumn{3}{l}{XML~\cite{lei2020tvr}}        &14.30 &5.58 &19.42 &17.27      \\
				\multicolumn{3}{l}{WSSL w/o subtitle} &2.07 &0.46 &2.52 &1.99      \\
				\multicolumn{3}{l}{WSSL~\cite{duan2018weakly} } &6.80 &2.91 &8.82  &7.01      \\
				\hline
				\multicolumn{3}{l}{DCIN w/o subtitle} &13.08  &4.51 &19.96 &16.63      \\
				\multicolumn{3}{l}{\textbf{DCIN-CCL}} &\textbf{19.21} &\textbf{7.16}  &\textbf{28.59}    &\textbf{22.73}   \\
				
				\bottomrule
			\end{tabular}
		\end{threeparttable}
	\caption{Performance comparison on the MovieGraph dataset.}
	\label{t1}
	\vspace{-0.5cm}
\end{table}

\section{Experiments}

\subsection{Experimental Setup}

\noindent
\textbf{Implementation Details.} For video, we use the ResNeXt-101 model~\cite{hara2018can} pre-trained on the kinetics-400 dataset as \cite{kukleva2020learning}. For subtitle, we employ a pre-trained BERT~\cite{devlin2018bert} and perform max pooling over each sentence to get the sentence representations. We set the dimension of segment and subtitle representations to 384. For the visual frames that are not aligned with any subtitles, we assign them to the neighboring segment-subtitle pair. For the hyper-parameters, we set $\Delta$ to 0.5, $\beta$ to 0.1, and set $\lambda_1, \lambda_2, \lambda_3, \lambda_4$ to 0.001, 1.0, 1.0, and 0.7, respectively. During training, we set the batch size to 32 and use Adam as optimizer~\cite{duchi2011adaptive}, where the learning rate is set to $1e^{-4}$.

\noindent
\textbf{Evaluation Metrics.} We employ \textbf{R@IoU}, \textbf{mIoU}, and \textbf{mAP} as evaluation metrics. The \textbf{R@IoU} is recall at various thresholds of the temporal Intersection over Union~(IoU). The R@IoU measures the percentage of predictions that have IoU with ground-truth larger than the thresholds. Here we set recall to 1, and temporal IoU threshold values to $\{0.5, 0.7\}$. \textbf{mAP} is the average precision over various IoU thresholds. \textbf{mIoU} is the average IoU between the predicted segments and ground-truth. For \textbf{mAP} and \textbf{mIoU}, we set temporal IoU threshold values to $\{0.1, 0.3, 0.5, 0.7\}$.

\noindent
\textbf{Baselines.} We compare the proposed approach with a number of strong baselines from relevant video-and-language tasks. Only publicly available models are used to calculate these metrics. Since the most related task with ours is action localization, we extend the existing weakly-supervised action localization~(WSAL) approaches \textbf{UntrimmedNets}~\cite{wang2017untrimmednets}, \textbf{3C-Net}~\cite{narayan20193c}, and \textbf{ASL}~\cite{ma2021weakly} as the baselines. Considering these baselines do not take subtitles as input, we implement two versions: 1) ignore subtitles directly; 2) fuse the subtitle features with aligned segment features. \textbf{UntrimmedNets} first learns a video-level classification and then selects frames with high classification activation as action locations. \textbf{3C-Net} adopts a classification loss to ensure the separability, a center loss to reduce inter-class variations, and a counting loss to delineate adjacent action sequences. \textbf{ASL} learns with a class-agnostic task to predict which frames will be selected by the classifier.

We also extend video-subtitle moment retrieval model \textbf{XML}~\cite{lei2020tvr} and weakly-supervised temporal grounding model \textbf{WSSL}~\cite{duan2018weakly} as baselines. \textbf{XML} is a recently proposed transformer-based method for TV show retrieval, which first encodes video and subtitle representation separately via two self-encoders, and then builds the cross-modality context representation via two cross-encoders. Here, we use the emotion label to attend to the above fused context features of videos and subtitles. To facilitate weakly-supervised learning, we adopt a multi-instance learning method~\cite{chen2020look, zhang2020counterfactual} to train the \textbf{XML} model. \textbf{WSSL} is a cycle system with a pair of dual problems: event captioning and sentence localization. Here, we train \textbf{WSSL} to reconstruct the emotion label as weakly-supervised objective. To show the importance of using both videos and subtitles, we compare baselines with their corresponding video-only variants and extend a standard  span-based QA model~\cite{huang2017fusionnet} as \textbf{subtitle-only} baseline. For \textbf{DCIN w/o subtitle}, we replace the cross-modal consensus learning with the weakly-supervised learning loss from \textbf{3C-Net}.

\begin{table}[!t]
	\centering
	\begin{tabular}{ ll|ccc}
		\hline
		&Method &R@0.5 &mAP &mIoU \\
		\hline
		
		1 &Backbone                                 &8.79    &12.36  &9.66\\
		2 &\quad + Coarse-Fine                &10.89   &16.39  &12.89 \\
		3 &\quad + TCDR                          &11.19    &17.64   &13.79 \\
		4 &\quad + GTCI (w/o CCF)           &12.99    &20.85  &16.62 \\
		5 &\quad + CCF = \textbf{DCIN}   &13.99	&22.02	&17.36    \\
		\hline
		
	\end{tabular}
	\caption{Performance comparison by varying the individual components of the DCIN.}
	\label{t2}
	\vspace{-0.9cm}
\end{table}

\vspace{-0.1cm}
\subsection{Results}
We compare our approach to the state-of-the-art WSAL methods, video-subtitle moment retrieval, and weakly-supervised temporal sentence grounding. We summarize the results in Table \ref{t1}. From the results, we can see that our method significantly outperforms the baselines, and the superiority is consistently observed on all metrics. We notice that our \textbf{DCIN w/o subtitle} also surpasses baselines on mAP, indicating the effectiveness of our DCIN on multi-granularity temporal context modeling. When using only subtitles to localize emotions, the span-based QA model only achieves 7.49\% on mAP. Also, we observe consistent improvement from subtitles on all baselines. These indicate the importance of multi-modal context.

 Furthermore, all adapted baselines from three relevant tasks perform poorly on fine-grained emotion localization. We speculate the main reasons are three folds: 1) As the actions and events have more consistent visual patterns, the methods in WSAL and temporal sentence grounding make predictions for each segment separately, ignoring the multi-granularity context. In contrast, we adaptively integrate different granularities of context into segment representations in a hierarchy. 2) The methods in WSAL and temporal sentence grounding are designed for localizing actions or events in pure videos without subtitles. They fail to effectively leverage the rich complementary information in subtitles. Differently, our approach utilizes the cross-modality consensus between video and subtitle to form an effective weakly-supervised learning paradigm. 3) Emotion classes have much higher inter-class similarity than action classes. The methods from all three tasks fail to learn discriminative context, which is crucial for inferring fine-grained information. In our approach, we propose a context-sensitive constraint to encourage the model to learn emotion-discriminative context.

\begin{table}[!t]
	\resizebox{\linewidth}{!}{
		\centering
		\begin{tabular}{ l|cc|ccc|ccc}
			\hline
			\multirow{2}*{}
			&\multicolumn{2}{c|}{TAR}
			&\multicolumn{3}{c|}{Loss Terms}   
			
			&\multirow{2}*{R@0.5}
			&\multirow{2}*{mAP}
			&\multirow{2}*{mIoU}
			\\
			
			&Hard &Relaxed &$\mathcal{L} _{cs}$ &$\mathcal{L} _{vc}$ &$\mathcal{L} _{sc}$ &\multicolumn{3}{c}{} \\
			
			\hline    
			
			1&\Checkmark  &      &    &   &   &13.99		&22.02	&17.36       \\
			
			2&  &\Checkmark     &    &  &   &14.99	&24.20	 &19.14        \\			
			
			\hline
			3& &\Checkmark      &\Checkmark     &  &      &16.55	&25.71	&20.29        \\
			4& &\Checkmark      &    &\Checkmark   &      &17.38	&26.48	&21.04        \\
			5& &\Checkmark      &    &   &\Checkmark      &16.25	&25.13	&19.84        \\		
			\hline				
			
			6& &\Checkmark   &\Checkmark &\Checkmark    &  &18.32	&27.40	&21.72      \\ 
			7& &\Checkmark  &\Checkmark & &\Checkmark      &16.75	&26.86	&21.20       \\ 
			8& &\Checkmark  & &\Checkmark &\Checkmark      &18.38	&27.62	&21.75       \\ 
			
			\hline
			9& &\Checkmark  &\Checkmark &\Checkmark &\Checkmark 
			&\textbf{19.21}  &\textbf{28.59}   &\textbf{22.73}     \\ 
			\hline
		\end{tabular}
		
	}
	\caption{Ablation studies of the temporal alignment relaxation and the proposed loss terms.}
	\label{t3}
\end{table}

\begin{figure}[!t]
	\centering
	\includegraphics[width=\linewidth]{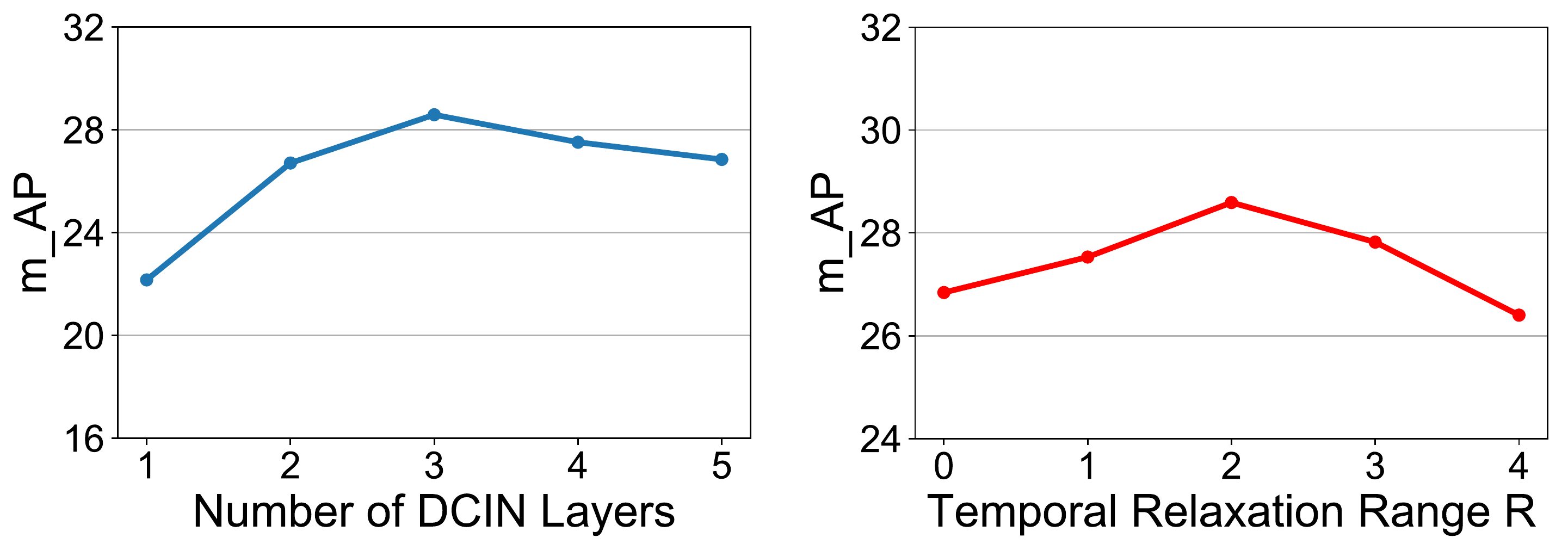}
	\vspace{-0.6cm}
	\caption{Ablation studies with respect to the number of DCIN layers and the temporal relaxation range.}
	\label{ablation}
\end{figure}

\subsection{In-Depth Analysis}

\noindent
\textbf{Effectiveness of Individual Component.} We first investigate the contribution of the Dilated Context Integrated Network~(DCIN) in Table \ref{t2}. We start with the backbone model and gradually add the Coarse-Fine architecture, Temporal Context Dependency Reasoning~(TCDR), Gated Temporal Context Integration~(GTCI), and cross-modal context filtering mechanism~(CCF) to form complete DCIN. To clearly distinguish the improvement, we only use the basic cross-modal consensus loss~($\mathcal{L}_{cmc}$) to train these ablation models without temporal relaxation and other proposed losses. As the transformer-based~\cite{vaswani2017attention} model is a powerful model that effectively captures no-local context, we use it as the backbone for context modeling. It takes video segment sequences $\{v_i\}^T_{i=1}$ and subtitle sentence sequences $\{s_i\}^T_{i=1}$ as input, performs self-attention and cross-modal attention on them, and finally outputs the context-aware segment and subtitle representations. 

As shown in Table \ref{t2}, the performance increases consistently, indicating the effectiveness of each component. Overall, our DCIN takes up 9.66\% of the gain on mAP. Particularly, the results from Row 2 to Row 4 suggest that our DCIN can better model the complex temporal context by adaptively integrating multi-granularity temporal context in a coarse-fine architecture. The results of Row 5 validate the superiority of the cross-modal context filtering mechanism, which utilizes the multi-modality context to better guide the context integration process. 


We then verify the strength of our temporal alignment relaxation~(TAR) and the proposed losses in Table \ref{t3}. We start with the complete DCIN~(Row 1). The results of Row 2 show that TAR improves the DCIN by dynamically learning the cross-modal alignment, which alleviates the misalignment noise during CCL. Furthermore, the results from Row 3 to Row 5 validate that each loss is helpful for emotion localization. Specifically, context-sensitive constraint~($\mathcal{L}_{cs}$) takes up 6\% of the relative gain on mAP and mIoU, the video consistency loss $\mathcal{L} _{vc}$ contributes 1.69\% and 1.43\% to the improvement on mAP and mIoU, respectively, and the subtitle consistency loss $\mathcal{L}_{sc}$ takes up 1.15\% and 0.91\% of gain on mAP and mIoU, respectively. In the end, the results from Row 6 to Row 9 suggest that the proposed losses can promote the cross-modal consensus and context sensitivity in a mutually rewarding way.

	\begin{table}[!t]
		\setlength\tabcolsep{4pt}
		\centering
		\begin{tabular}{ ll|ccc|ccc}
			\hline
			\multicolumn{2}{l|}{\multirow{2}*{Method}} 
			&\multicolumn{3}{c|}{THUMOS-14} 
			&\multicolumn{3}{c}{ActivityNet Captions}
			
			\\
			\multicolumn{2}{l|}{}  &R@0.3  &R@0.5  &\multicolumn{1}{c|}{R@0.7}  &R@0.3 &R@0.5  &\multicolumn{1}{c}{mIoU}   \\
			\hline    
			\multicolumn{2}{l|}{3C-Net \cite{narayan20193c}}    &40.9   &24.6    &7.7        &-   &-   &-   \\
			\multicolumn{2}{l|}{TSCN \cite{zhai2020two}}          &47.8   &28.7    &10.2        &-   &-   &-   \\
			\multicolumn{2}{l|}{ASL \cite{ma2021weakly}}           &51.8   &31.1    &11.4         &-   &-   &-  	  \\
			\hline
			\multicolumn{2}{l|}{WSSL \cite{duan2018weakly}}  &-   &-   &-      &41.98    &23.34     &28.23     \\
			\multicolumn{2}{l|}{SCN \cite{lin2020weakly}}  &-   &-   &-      &47.23    &29.22     &-     \\
			\multicolumn{2}{l|}{VGN \cite{zhang2020counterfactual}}      &-   &-   &-      &50.12    &31.07     &-        \\
			\hline
			\multicolumn{2}{l|}{DCIN}    &50.3    &29.8    &11.9       &46.72    &28.19     &33.74       \\
			\hline
			
		\end{tabular}
		\caption{Transferability of our DCIN.}
		\label{t4}
	\end{table}

\begin{figure*}
	\begin{center}
		\includegraphics[width=\textwidth]{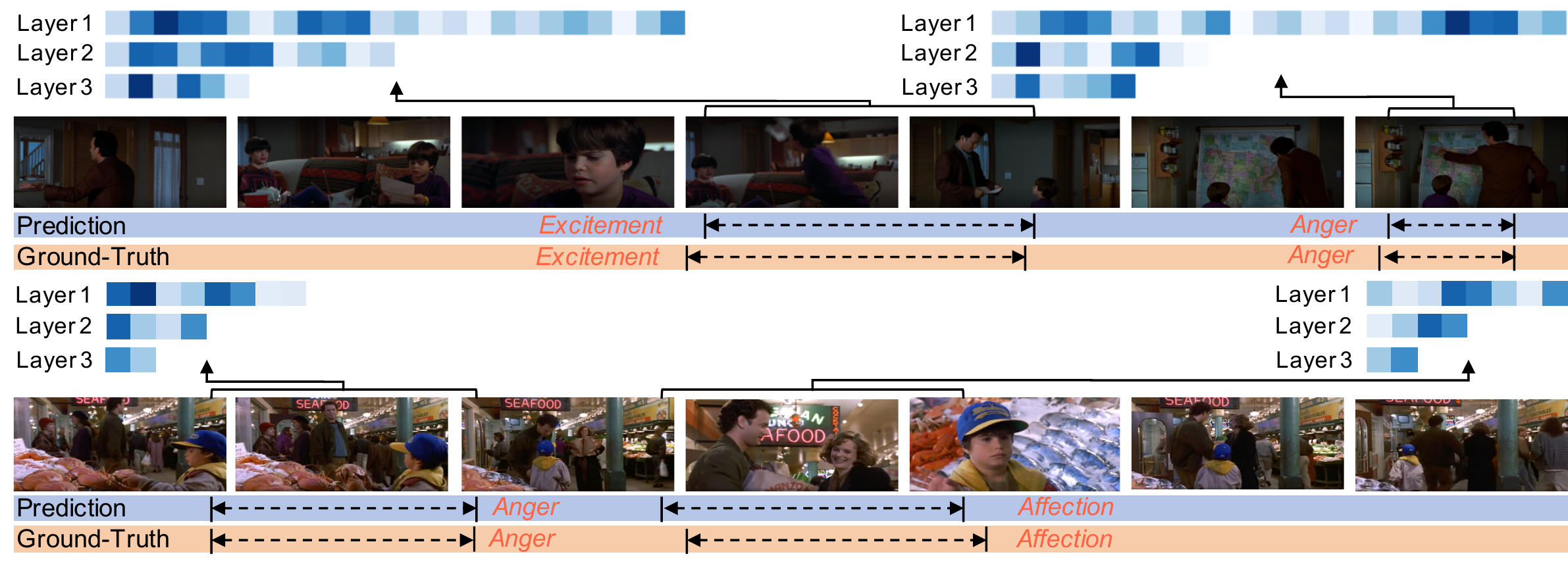}
	\end{center}
	\caption{Qualitative examples of our proposed model.} 
	\label{qualitative}
\end{figure*}

\begin{figure}[!t]
	\centering
	\includegraphics[width=\linewidth]{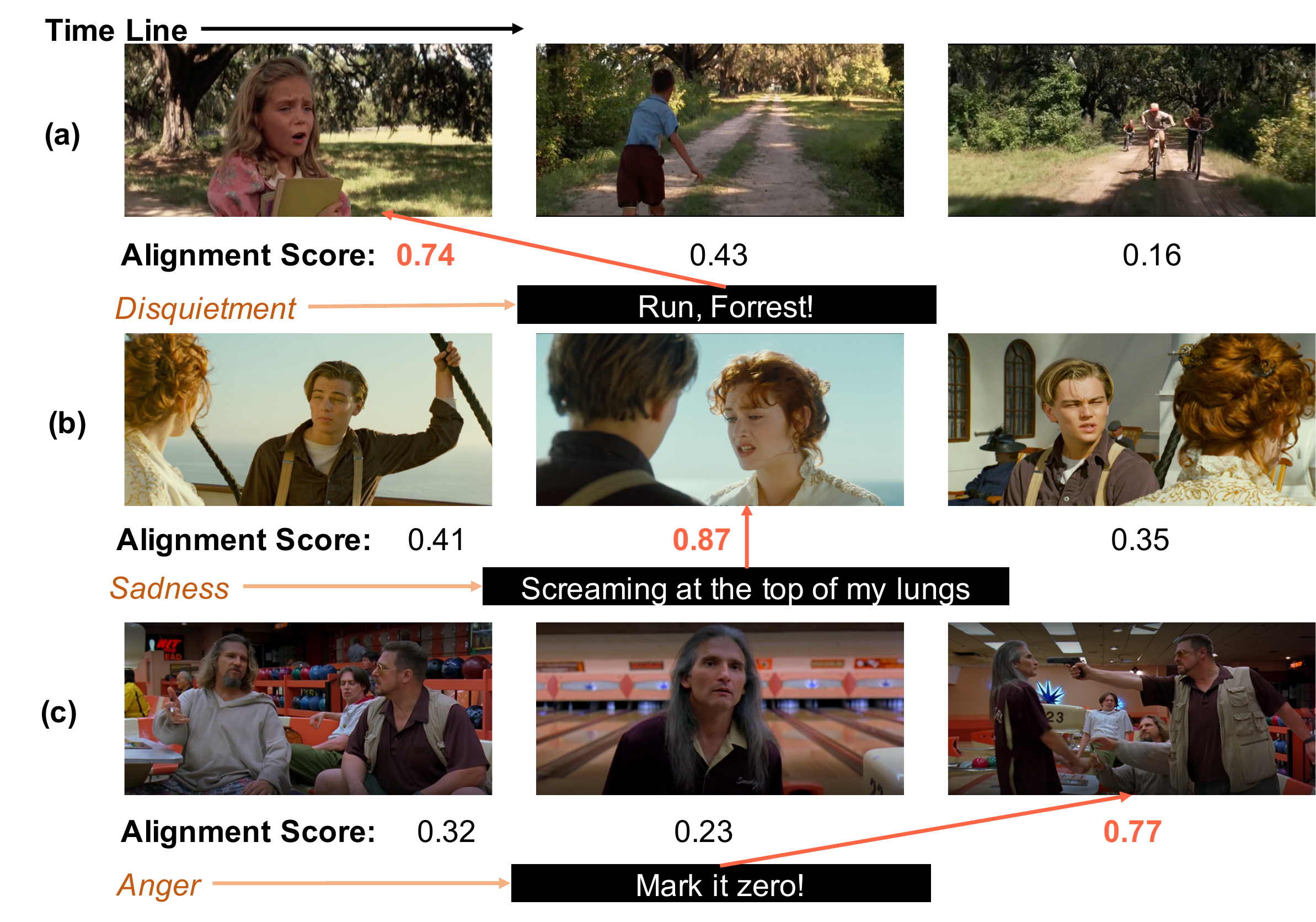}
	\caption{Qualitative examples of CCL. The middle column shows the most relevant subtitles of the given emotions and the temporally aligned video segments.}
	\label{case_ccl}
\end{figure}

\noindent
\textbf{Impact of DCIN Layer Numbers.} Figure \ref{ablation} presents the results across the number of DCIN layers, where performance increases until 3 layers and decreases afterward. This indicates that 3 DCIN layers can capture enough granularities of temporal context for fine-grained emotion understanding.

\noindent
\textbf{Analysis on Temporal Alignment Relaxation.} We explore the impact of temporal relaxation range, where $R$ means that we consider the $R$ nearest subtitle~(segment) to the central subtitles~(segments) in each side as candidates. $R=0$ corresponds to the hard temporal alignment version.
As shown in Figure \ref{ablation}, the performance keeps increasing when the $R$ is increased from 0 to 2. When we continue to increase the $R$, too many candidates introduce larger noise. 

\subsection{Transferability to Different Tasks}

We further evaluate DCIN on temporal action localization and temporal sentence grounding tasks to illustrate its superiority on multi-granularity temporal context modeling. For  temporal action localization, we add fully-connected layers to predict the confidence score of each segment feature encoded by DCIN, and directly adopt the weakly-supervised learning paradigm from 3C-Net to train DCIN. For temporal sentence grounding, we first adopt our DCIN to obtain context-aware segment representations, then generate multiple proposals and fuse proposals with query sentences as 2D-TAN~\cite{zhang2020learning}, and finally use a multi-instance learning objective to train DCIN.

Table \ref{t4} summarizes the temporal action localization performance on THUMOS-14~\cite{THUMOS15} dataset and the temporal sentence grounding performance on ActivityNet Captions~\cite{krishna2017dense} dataset. We notice that the coarse-fine two stream architecture for adaptively multi-granularity temporal dynamics reasoning can also benefit the accurate temporal action localization, improving R@0.7 from 11.4\% to 11.9\%. Meanwhile, our DCIN can achieve comparable performance on temporal sentence grounding task. 


\subsection{Qualitative Analysis}

\noindent
\textbf{Case Study and Visualization of DCIN.} Figure \ref{qualitative} visualizes two qualitative examples. Evidently, our model can produce accurate temporal boundaries for the TEL task. For a more intuitive view of how DCIN adaptively integrates multi-granularity context, we also visualize the context gate values of the target segments at different DCIN layers, which reflect the context that the target segments focus on. The context at different time scales is produced by the Coarse stream at different DCIN layers. For instance, for the segment corresponding to emotion \textsl{Excitement}~(Figure \ref{qualitative}.a), its context gates are well activated for the context at all layers that contains the information of the boy happily receiving a letter.

\noindent
\textbf{Visualization of CCL.} In Figure \ref{case_ccl}, we show three examples of how the CCL computes the semantic alignment between subtitle and video. In (a), the subtitle at the second column is the most relevant subtitle for emotion \textsl{disquietment} based on $P(e|s_i)$, and the segment at the second column is the temporally aligned segment. The left column and right column correspond to the previous and forthcoming segments, respectively. Below the segments are the semantic alignment scores between them and the subtitle. We can see that the subtitle is said by the girl in the previous segment. If we only follow the hard temporal alignment relationship, we will optimize the segment at the middle column to predict high a score for emotion \textsl{disquietment}, which might confuse the model. By considering the temporal alignment relaxation on a neighboring window of it, our CCL adaptively matches the semantic aligned segment~(the left column) based on the learned semantic alignment scores. Moreover, the second example shows the case that the temporal alignment is consistent with the semantic alignment, and the third example shows the case that the subtitle refers to the forthcoming segment. These cases indicate that our CCL can handle the misalignment noisy and dynamically learns the cross-modal semantic alignment.

\section{Conclusions}
In this paper, we define a novel task of temporal emotion localization, which fosters deeper investigations in emotion understanding and video-and-language reasoning. To solve the challenges in the task, we propose a novel dilated context integrated network to adaptively integrate multi-granularity temporal context in a hierarchy, as well as a cross-modal consensus learning paradigm for weakly-supervised learning. The experimental results show the effectiveness and transferability of the proposed framework.

\section*{Acknowledgment}
This work has been supported in part by National Key Research and Development Program of China (2018AAA0101900), Zhejiang NSF (LR21F020004), Alibaba-Zhejiang University Joint Research Institute of Frontier Technologies, Key Research and Development Program of Zhejiang Province, China (No. 2021C01013), Chinese Knowledge Center of Engineering Science and Technology (CKCEST). We thank all the reviewers for valuable comments.

\bibliographystyle{ACM-Reference-Format}
\balance
\bibliography{sample-sigconf}

\end{document}